# ARCHITECTURE OF A FUZZY EXPERT SYSTEM USED FOR DYSLALIC CHILDREN THERAPY

*assistant drd. eng., SCHIPOR Ovidiu-Andrei,*
*professor PhD., PENTIUC Ştefan-Gheorghe,*
*lecturer PhD., SCHIPOR Maria-Doina*

*"Ştefan cel Mare" University, Suceava, Romania,*
*schipor@eed.usv.ro, pentiuc@eed.usv.ro, vmdoina@yahoo.com*

**Abstract**
*In this paper we present architecture of a fuzzy expert system used for therapy of dyslalic children. With fuzzy approach we can create a better model for speech therapist decisions. A software interface was developed for validation of the system.*

*The main objectives of this task are: personalized therapy (the therapy must be in according with child's problems level, context and possibilities), speech therapist assistant (the expert system offer some suggestion regarding what exercises are better for a specific moment and from a specific child), (self) teaching (when system's conclusion is different that speech therapist's conclusion the last one must have the knowledge base change possibility).*

## 1. Introduction

In this article we refer to LOGOMON system developed in TERAPERS project by the authors. The full system is used for personalized therapy of dyslalia affecting pre scholars (children with age between 4 and 7). Dyslalia is a speech disorder that affect pronunciation of one ore many sounds. According to the statistics, about 10% of pre scholars are affected by this type of speech impairment [1].

The objectives of LOGOMON system are:
- initial and during therapy evaluation of volunteer children and identification of a modality of standardizing their progresses and regresses (physiological and behavioral parameters);

- rigorous formalization of an evaluation methodology and development of a pertinent database for the speech productions;
- the development of an expert system for the personalized therapy of speech impairments that will allow for designing a training path for pronunciation, individualized according to the defect category, previous experience and the child's therapy previous evolution;
- the development of a therapeutically guide that will allow mixing classical methods with the adjuvant procedures of the audio visual system and the design of a database that will contain the set of exercises and the results obtained by the child.

All these activities are currently completed and the therapy system is tested by Interschool Regional Speech Therapy Center of Suceava, Romania.

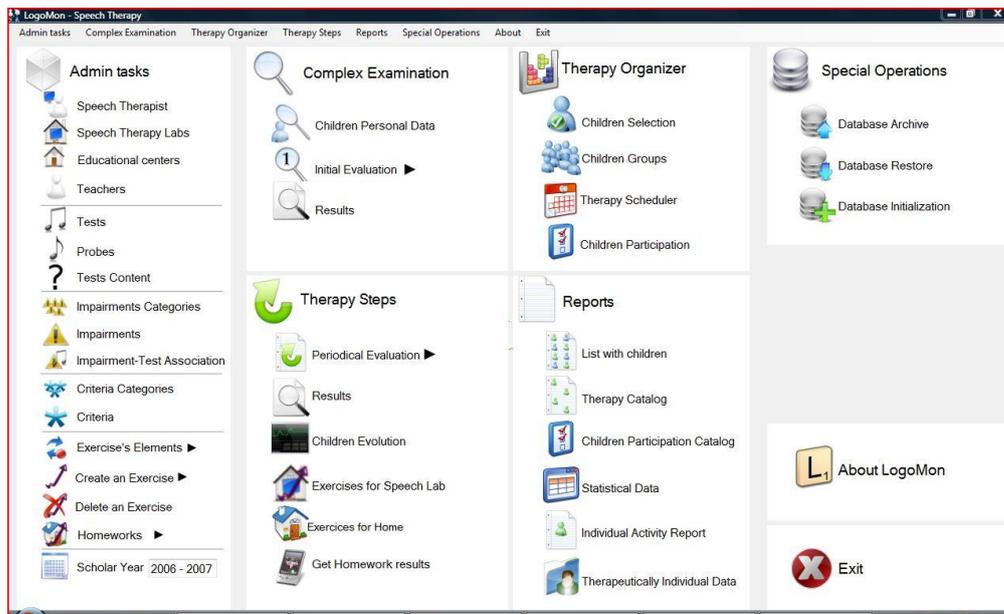

*Figure 1 - LOGOMON speech therapy system*

The figure 1 presents the menus of LOGOMON (Administrative Tasks, Complex Examination, Therapy Organizer, Therapy Steps, Reports, Special Operations, About, Exit).

## 2. Literature review

There is a powerful preoccupation at the European level in helping the people with such disorders because of highly social and affective implications.

Forward we enumerate most important projects related with speech therapy area:
- OLP (Ortho-Logo-Paedia) - develop in „*EU Quality of Life and Management of Living Resources*" interest area with coordination of *Institute for Language and Speech Processing*, Athena, Greek, with participation of France (*Arches*), Greek (Logos Centre for Speech-Voice Pathology), Spain (Universidad *Politecnica de Madrid*), Sweden (*KTH- Royal Institute of Technology*) and Great Britain (*Sheffield University, Barnsley District General Hospital*) [2], [3];
- ARTUR (ARticulation TUtoR) – one of the most recently speech therapy system [4], [5], still develop in 2006 year, with coordination of *KTH-Royal Institute of Technology*, Sweden;
- STAR (Speech Training, Assessment and Remediation) – a system develop by *Alfred I. duPont Hospital for Children* and *University of Delaware* for help speech therapist and children with speech problems [6];
- MMTTS-CSD (Multimedia, Multilingual Teaching and Training System for Children with Speech Disorders) – a complex project develop by *University of Reading – Anglia, Budapest University of Technology and Economics – Ungaria, University of Maribor – Slovenia* and *Kungl. Tekniska Hogskolan,* Sweden [7].

At the national level, little research has been conducted on the therapy of speech impairments, out of which mostly is focused on traditional areas such as voice recognition, voice synthesis and voice authentication. Although there are a lot of children with speech disorder, the methods used today in logopaedia are mostly based on individual work with each child. The few existent computer assisted programs in Romania don't provide any feedback. At international level, there are software applications but quite expensive and improper for the phonetic specific of Romanian language.

## 3. Logo-therapeutic aspects related personalized therapy

In figure 2 are presented the main aspects related preschools dyslalia therapy. In complex examination, speech therapist collects the base set of data used for child diagnostic.

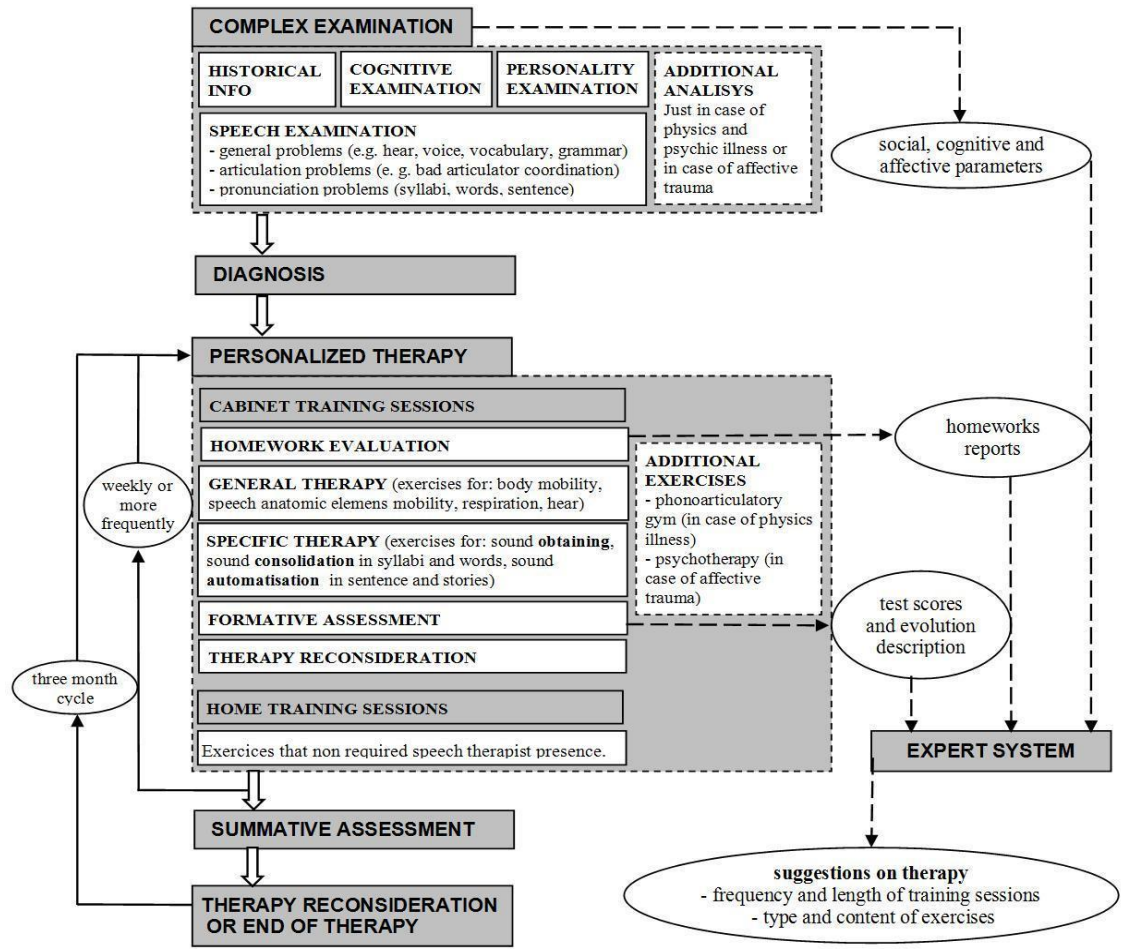

*Figure 2 - Personalized therapy steps and role of expert system*

Speech examination has two distinct levels [8]:
- examination of hear, voice, grammatical and linguistic skills;
- speech production (articulator system and vocal production).

Therapeutically program has two steps [9]:
- generally therapy (mobility development exercises, air flow control, hear development);
- specific therapy (sound pronunciation, consolidation and differentiation).

In according with [1], speech therapy software can help speech problems diagnostic, can offer a real-time, audio-visual feedback, can improve analyze of child progress and can extend speech therapy at child home.

## 4. Expert system – fuzzy approach

With fuzzy approach we can create a better model for speech therapist decisions. A software interface was developed for validation of the system. The main objectives of this task are:
- personalized therapy (the therapy must be in according with child's problems level, context and possibilities);
- speech therapist assistant (the expert system offer some suggestion regarding what exercises are better for a specific moment and from a specific child);
- (self)teaching (when system's conclusion is different that speech therapist's conclusion the last one must have the knowledge base change possibility).

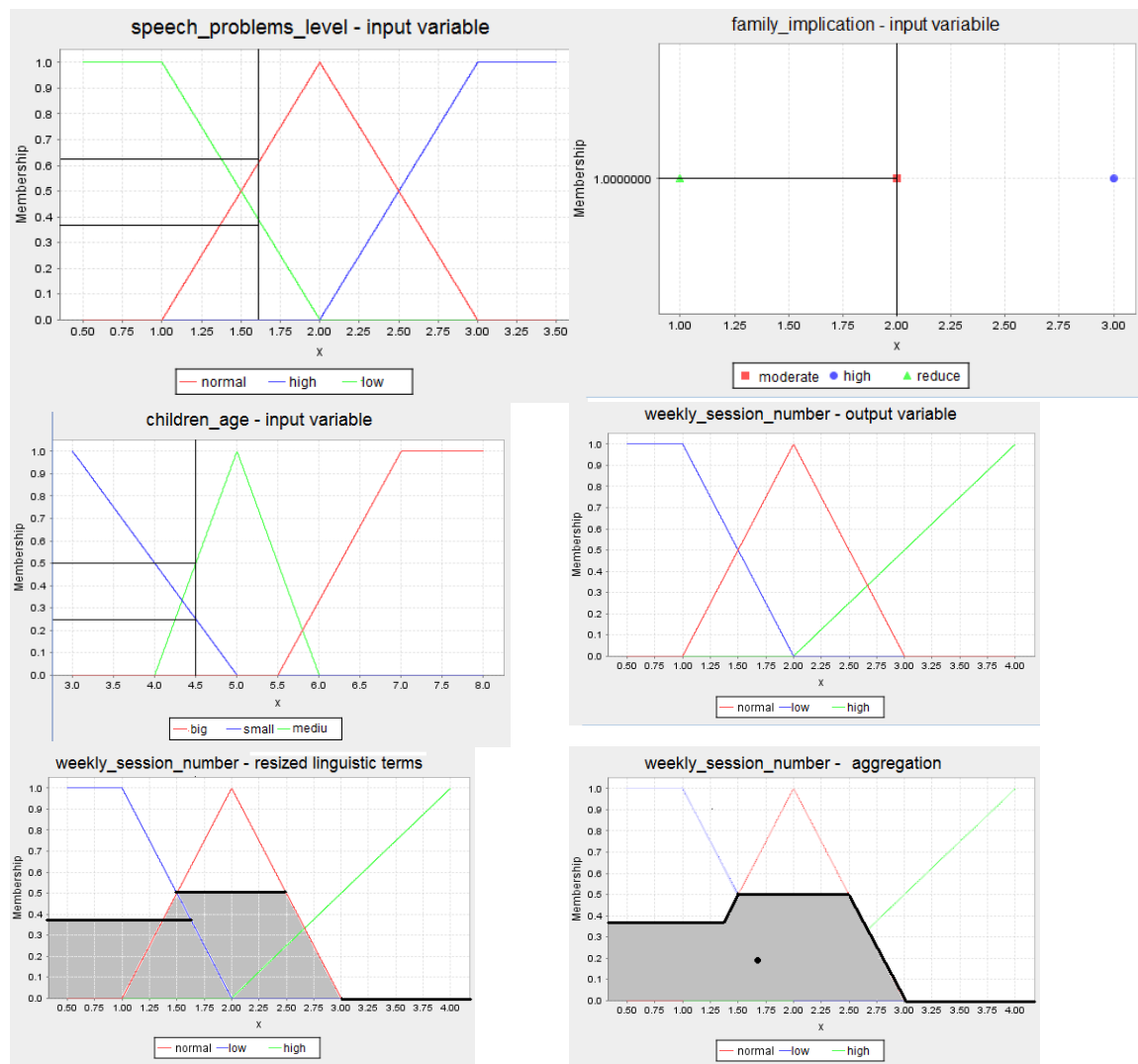

*Figure 3 - Example of speech therapy fuzzy inference*

Fuzzy logic has ability to create accurate models of reality. It's not an "imprecise logic". It's a logic that can manipulate imprecise aspects of reality. In the latest years, many fuzzy expert systems were developed [10], [11]. In figure 3 we present an example of fuzzy inference, using three input linguistic variables (speech problems level, family implication and children age) and one output linguistic variable (weekly session number).

First three variables have following textual representation:
*speech_problems_level (1.62) = {"low"/0.37,"normal"/0.62,"high"/0.0}*
*family_implication (2.00) = {"reduce"/0.0,"moderate"/1.0,"high"/0.0}*
*children_age (4.50) = {"small"/0.25,"medium"/0.5,"big"/0.0}*

We consider five rules for illustrate the inference steps:

- *IF (speech_problems_level is high) and (child_age is medium) and (family_implication is reduce) THEN weekly_session_number is high;* min (0.00, 0.50, 0.00) = **0.00** for linguistic term **high**

- *IF (speech_problems_level is low) and (child_age is small) and (family_implication is moderate) THEN weekly_session_number is low;* min (0.37, 0.25, 1.00) = **0.25** for linguistic term **low**

- *IF (speech_problems_level is low) and (child_age is medium) and (family_implication is moderate) THEN weekly_session_number is low;* min (0.37, 0.50, 1.00) = **0.37** for linguistic term **low**

- *IF (speech_problems_level is normal) and (child_age is small) and (family_implication is moderate) THEN weekly_session_number is normal* min (0.62, 0.25, 1.00) = **0.25** for linguistic term **normal**

- *IF (speech_problems_level is normal) and (child_age is medium) and (family_implication is moderate*

THEN *weekly_session_number is normal*
min (0.62, 0.5, 1.00) = **0.50** for linguistic term **normal**

Final coefficients levels are obtained using max function:
**high** = *max (0.00)* = **0.00**;
**low** = *max (0.25, 0.37)* = **0.37**;
**normal** = *max (0.25, 0.50)* = **0.50**

Each linguistic term of output variable has another representation and in this manner is obtained final graphical representation of weekly_session_number variable. If system user wants to get a single output value, then area center of gravity is calculated. In our case

(value 1.62), child must participate at one to two session (but two is preferred).

We implement over 150 fuzzy rules for control various aspects of personalized therapy. These rules are currently validated by speech therapists and can be modified in a distributed manner. We also develop a Java expert engine interface in order to test our knowledge base [9], [12].

### Acknowledgements

We must specify that these researches are part of TERAPERS project financed by the National Agency for Scientific Research, Romania, INFOSOC program (contract number: 56-CEEX-II-03/27.07.2006).